\pdfoutput=1

\documentclass[11pt]{article}

\usepackage[final]{acl}
\usepackage{graphicx}
\usepackage{booktabs}
\usepackage{array}
\usepackage{tabularx}
\usepackage{adjustbox}
\usepackage{subcaption}
\usepackage{times}
\usepackage{latexsym}

\usepackage{xcolor}
\usepackage{amssymb}  
\usepackage{amsmath}

\usepackage[T1]{fontenc}

\usepackage[utf8]{inputenc}

\usepackage{microtype}
\usepackage{amsmath}
\usepackage{float}
\usepackage{inconsolata}

\usepackage{graphicx}
\usepackage{multicol}
\usepackage{longtable}
\usepackage{multirow}
\usepackage{booktabs} 
\usepackage{array} 
\usepackage{comment}

\title{The Geometry of Numerical Reasoning:\\ Language Models Compare Numeric Properties in Linear Subspaces}

\author{
\normalsize Ahmed Oumar El-Shangiti$^{1}$ \hspace{5mm} Tatsuya Hiraoka$^{1}$
 \hspace{5mm} \textbf{Hilal AlQuabeh}$^{1}$ \\ \textbf{Benjamin Heinzerling}$^{3, 2}$
\normalsize \hspace{5mm} \textbf{Kentaro Inui}$^{1, 2, 3}$ \\
\normalsize $^{1}$ Mohamed bin Zayed University of Artificial Intelligence (MBZUAI) \\
\normalsize $^{2}$Tohoku University \\
\normalsize $^{3}$RIKEN \\
\normalsize \texttt{ahmed.oumar@mbzuai.ac.ae}
}

\begin{document}
\maketitle
\begin{abstract}
This paper investigates whether large language models (LLMs) utilize numerical attributes encoded in a low-dimensional subspace of the embedding space when answering questions involving numeric comparisons, e.g., \textit{Was Cristiano born before Messi?}.
We first identified, using partial least squares regression,  these subspaces, which effectively encode the numerical attributes associated with the entities in comparison prompts. 
Further, we demonstrate causality, by intervening in these subspaces to manipulate hidden states, thereby altering the LLM's comparison outcomes. 
Experiments conducted on three different LLMs showed that our results hold across different numerical attributes, indicating that LLMs utilize the linearly encoded information for numerical reasoning.
\end{abstract}

\section{Introduction}
Language models (LMs) store large amounts of world knowledge in their parameters \cite{petroni2019language,jiang2020how,roberts2020much,heinzerling2021language,kassner2021multilingual}.
While prior work has evaluated parametric knowledge mainly via behavioral benchmarks, more recent work has analyzed how knowledge is represented in activation space, for example, localizing relational knowledge to specific layers and token representations \cite{meng2022locating,geva2023dissecting,merullo2024a} or identifying subspaces that encode numeric properties such as an entity's birth year \cite{heinzerling2024monotonicrepresentationnumericproperties}.
However, analysis of LM-internal knowledge representation has been limited to simple factual recall, e.g., for queries like ``When was Cristiano born?'' (Answer: 1985) or ``When was Messi born?'' (Answer: 1987).
If and how the mechanisms responsible for simple factual recall also participate in more complex queries, e.g., ``Is Cristiano older than Messi?'', is not understood so far.
A possible mechanism by which an LLM answers this query is a multi-step process consisting of first recalling the respective birth years of the two entities, comparing the two years, and then selecting a corresponding answer.


Herein, we focus on LLM's ability of arithmetic operations~\cite{dehaene2011number}.
The LLM's ability to handle numbers has been discussed after the advent of pre-trained language models~\cite{spithourakis2018numeracy,wallace2019nlp}.
With modern LLMs such as the LLaMA family~\cite{touvron2023llama2}, \newcite{heinzerling2024monotonicrepresentationnumericproperties} shows that LLMs map numerical attributes such as \textit{(Cristiano, born-in, 1985)} and \textit{(Messi, born-in, 1987)} to low-dimensional (Linear) subspaces and prove that those subspaces are used during knowledge extraction. 
However, it is not clear whether the LLMs use those subspaces to solve logical reasoning such as the relation \textit{(Cristiano, born-before, Messi)}.

\begin{figure}[t]
    \centering
    \includegraphics[width=\linewidth]{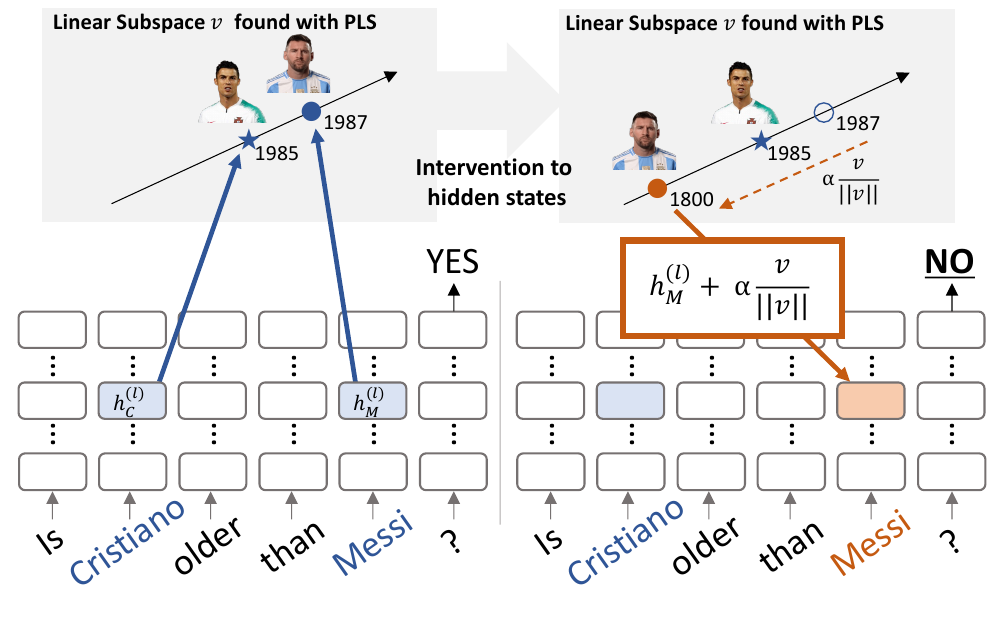} 
    \caption{Summary of our approach. We extract contextualized numeric attribute activations and then train $k$-components PLS model on the activations to predict their values and then use the first component of the PLS model to do an intervention at the last token of the second entity in the logical comparison. }
    \label{fig:figure1(summary of our approach)}
\end{figure}

In this study, we tackle the research question: \textbf{do LLMs leverage the linear subspace of entity-numerical attributes when solving numerical reasoning tasks?}
We investigate whether the linear subspace is indeed used in the logical reasoning tasks.
We first show the LLMs' capability to solve the numerical reasoning tasks from the viewpoint of behavioral observation: testing the performance of the reasoning task with in-context learning (\S \ref{Behavioral Experiments}).
We then examine the representations of LLMs (\S \ref{Internal Representation Experiments}).
We identify the linear subspace corresponding to the numerical attributes with partial least-squares (PLS~\cite{pls}) and intervene in the representation to test whether the model utilizes the linearly represented information (see Figure \ref{fig:figure1(summary of our approach)}).

The experimental results on the three numerical properties (the birth/death year of a person and the latitude of location) and on three LLMs (LLama3 8B~\cite{dubey2024llama3herdmodels}, Mistral 7B~\cite{jiang2023mistral7b}, and Qwen2.5 7B~\cite{qwen2.5} all instruction based models) demonstrate that LLMs leverage the numerical information represented in the linear subspace for the reasoning tasks.

\section{Outline of Experiments}
\begin{table}
\centering
\small
\begin{tabular}{@{}p{0.22\columnwidth}p{0.48\columnwidth}p{0.22\columnwidth}@{}}
\toprule
\textbf{Experiment} & \textbf{Question} & \textbf{Response} \\
\midrule
\multirow{3}{*}{\textbf{Extraction}} & Birth year of Albert Einstein? & 1879 \\
 & What is Isaac Newton's year of death? & 1727 \\
 & Latitude of Cairo? & 30.04° N \\
\addlinespace
\multirow{3}{*}{\textbf{Reasoning}} & Einstein born before Newton? & No \\
 & Einstein died before Newton? & No \\
 & Is Cairo's latitude higher than Jerusalem's? & Yes \\
\bottomrule
\end{tabular}
\caption{Samples from Extracting Information and Comparisons Experiments}
\label{table:experiments}
\end{table}

This section outlines our methodology to investigate the process of LLMs to solve the numerical reasoning.

\subsection{Model and Dataset}
In this work, we focus on the three numerical properties: the birth years of person entities, the death years of person entities, and the latitudes of location entities.
Table \ref{table:experiments} exemplifies the questions and expected responses for both tasks.
For the knowledge extraction task, we create the question-answer pairs by extracting 5,000 entities alongside their numerical attributes from Wikidata~\cite{helloworld}.
After filtering out entities that the LLM does not know (\S \ref{behavioraol_experiment_1}), we created the 5,000 questions about numerical reasoning that include two entities each.
For all experiments, we used Llama3-8B-instruction following model~\citep{dubey2024llama3herdmodels} as the LLM and later validate our finding on two additional models (see \S~\ref{subsection: Experiments on Additional Models}).
\subsection{Design of Experiments}
We conducted the experiments in two phases to investigate the LLM's ability to utilize the linear subspace for numerical reasoning.
\paragraph{Data Pre-processing (\S \ref{Behavioral Experiments}):} 
We began by evaluating the LLM's ability to handle both knowledge extraction and numerical reasoning tasks by inputting questions and evaluating its response.
To focus the subsequent experiments on entities for which the LLM has reliable numerical knowledge, we filtered out any entities that the LLM could not answer correctly during this initial behavioral experiment.
\paragraph{Internal Representation Experiments (\S \ref{Internal Representation Experiments}):}
In the second phase, we examined the inner workings of the LLM when solving the knowledge extraction (\S \ref{internal_representation_exp1}) and the numerical reasoning (\S \ref{internal_representation_exp2}). 
Here, we focus on analyzing the hidden state of each entity representation at a particular layer for knowledge extraction.
For the case of numerical reasoning, we investigated the activations of the last token's representation.
We denote the hidden state of the $i$-th input at the $l$-th layer as $h^{(l)}_i$. To investigate whether knowledge of numerical attributes is stored in low-dimensional subspaces, we applied PLS~\cite{pls} for each representation~\cite{heinzerling2024monotonicrepresentationnumericproperties}.
Partial Least Squares (PLS) offers an alternative to Principal Component Analysis (PCA) for dimensionality reduction, especially when predicting one set of variables from another. PLS seeks to maximize the covariance between the input matrix $\textbf{X}$ and the response matrix $\textbf{Y}$ by projecting both onto a latent space. Through PLS, we identified components that represent the linear structure of each numerical attribute, allowing us to analyze how the LLM might utilize these subspaces for reasoning.
To further test this, we intervened in the hidden state $h^{(l)}_i$ by incorporating the 1st PLS component $v$, as follows:
\begin{equation}
    \label{equation1}
    h^{(l)}_i \leftarrow h^{(l)}_i + \alpha \frac{{v}}{\|{v}\|},
\end{equation}
where $\alpha$ is a hyperparameter derived from the first PLS component, and $||v||$ is the Euclidian norm (L2-norm) of the vector $v$. 
Intuitively, this intervention edits the numerical attribute captured by the LLM.
For instance, if the numerical information \textit{(Cristiano, born-in, 1985)} is shifted to \textit{(Cristiano, born-in, 2020)}, an LLM that genuinely relies on a linear subspace for reasoning would adjust its interpretation accordingly, reflecting the change in its responses (Figure \ref{fig:figure1(summary of our approach)}).




\section{Data Pre-processing}
\label{Behavioral Experiments}
The purpose of this experiment is to assess whether the LLM possesses knowledge of the numerical attributes of the entities prepared for this study, and to evaluate its capability to perform numerical reasoning tasks.
Additionally, by conducting behavioral experiments focused on information extractions, we aim to filter out entities for which LLM lacks sufficient knowledge, therefore creating a refined dataset to be used in the subsequent numerical reasoning tasks.
For both tasks, extraction and reasoning, we prepared ten distinct prompts. The prompts that demonstrated the best performance in preliminary tests were selected for further investigation of the internal representations (\S \ref{Internal Representation Experiments}).
Appendix~\ref{tab:comprehensive_list_of_prompts} lists the complete list of prompts in the experiments.

\subsection{Knowledge Extraction}
\label{behavioraol_experiment_1}
To assess the LLM's knowledge extraction of entity numerical attributes, we conducted a zero-shot question-answering task, in which we asked direct questions about numerical attributes for various entities.
The results summarized in the top half of Table \ref{table:behavioral experiments results}, demonstrate that the LLM correctly answered at least 67\% of the prepared questions with the best-performing prompt for each task.
\subsection{Numerical Reasoning}
\label{behavioral_experiments_2}

For the numerical reasoning task, we created 5,000 question samples using a pair of unique entities, selected after filtering out those that the LLM could not answer correctly in \S \ref{behavioraol_experiment_1}.
Each question was designed to prompt the model to perform numerical reasoning, with binary (Yes/No) answers indicating correctness. The results, shown in the bottom half of Table \ref{table:behavioral experiments results}, reveal varying levels of accuracy across different prompts.
The LLM achieved around 75\% for birth/death year prediction, but only 56\% for latitude-related questions, suggesting differences in task difficulty.

\section{Internal Representation Experiments}
\label{Internal Representation Experiments}

\begin{figure*}[ht!]
    \centering
    \begin{subfigure}[b]{0.32\textwidth}
        \includegraphics[width=\textwidth]{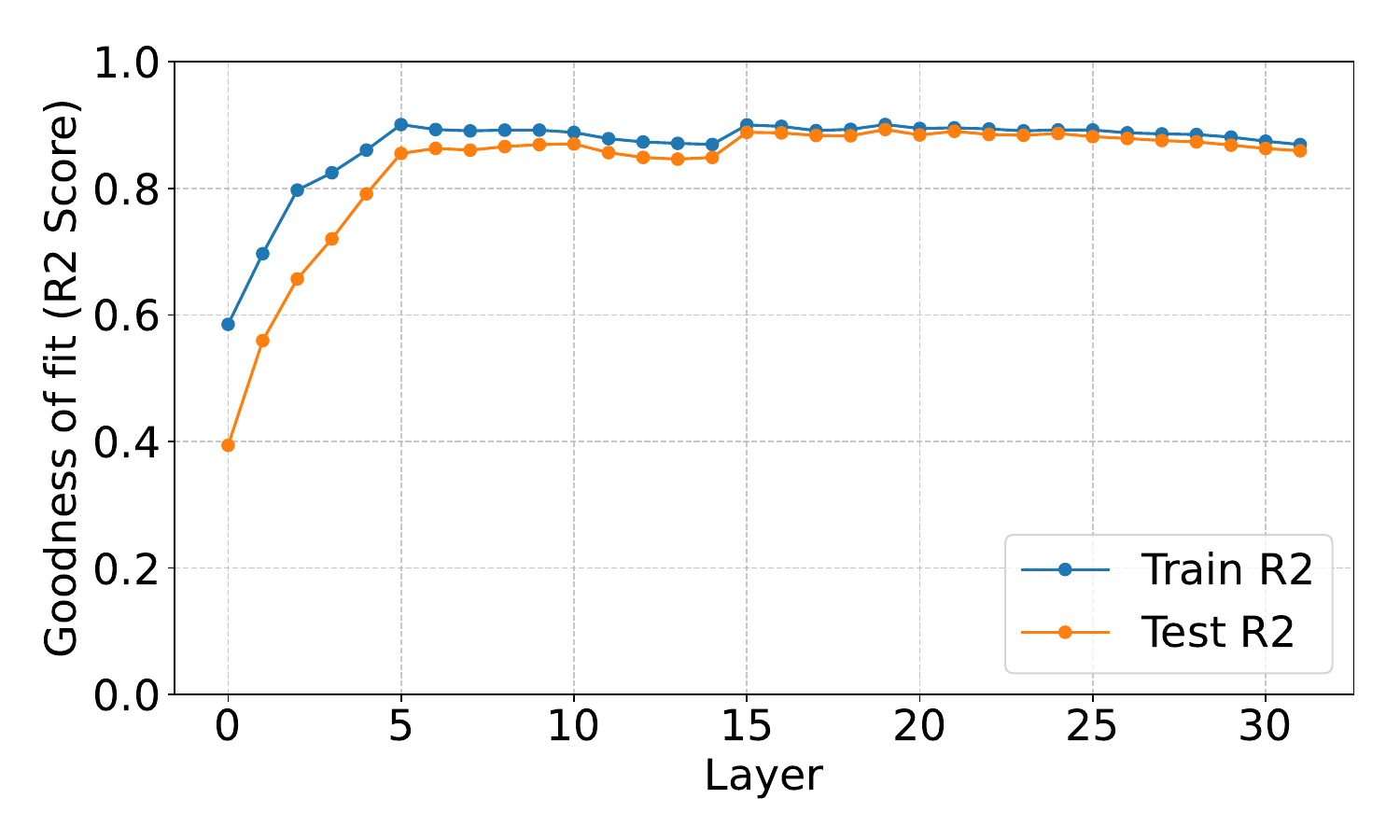}
        \caption{Birth Attribute}
    \end{subfigure}
    \hfill
    \begin{subfigure}[b]{0.32\textwidth}
        \includegraphics[width=\textwidth]{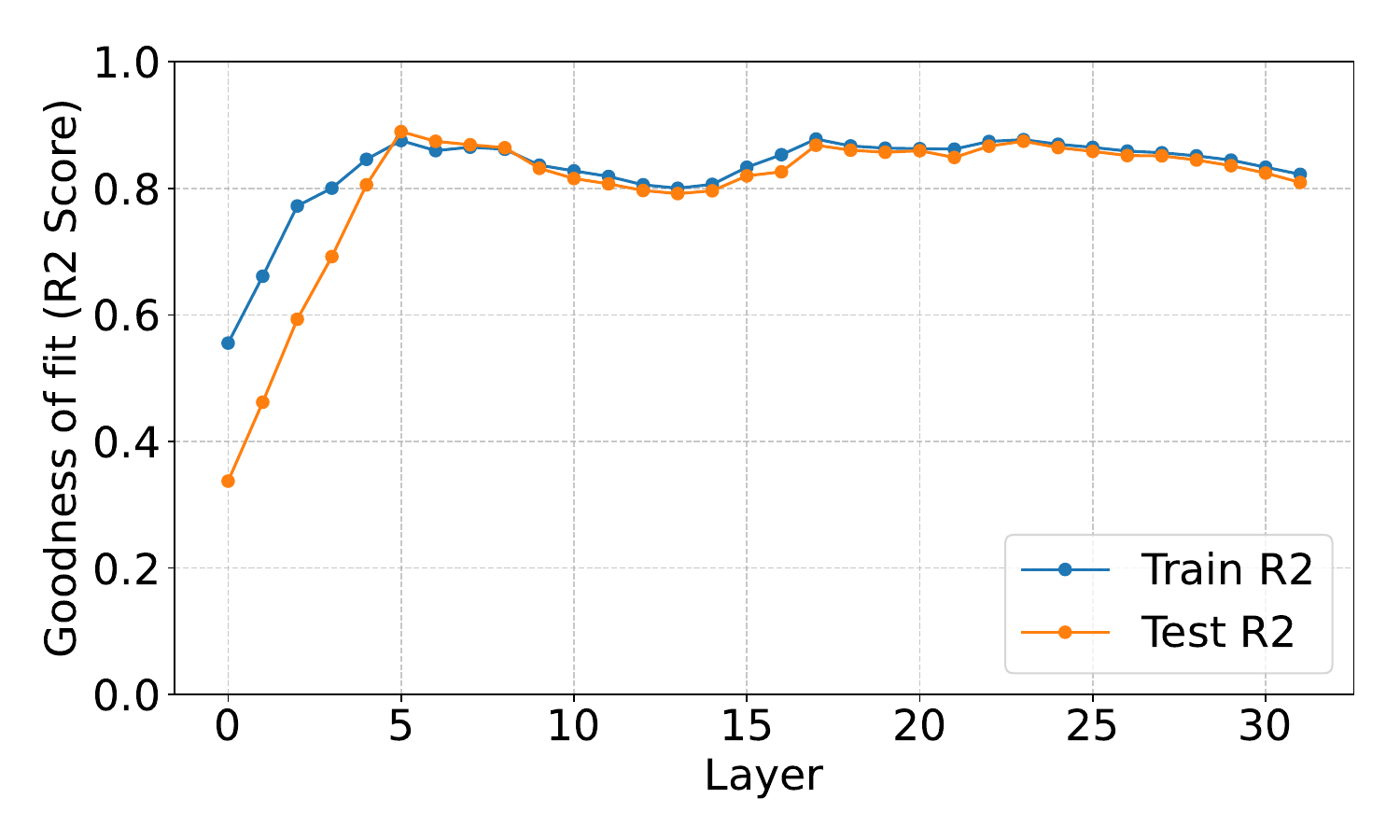}
        \caption{Death Attribute}
    \end{subfigure}
    \hfill
    \begin{subfigure}[b]{0.32\textwidth}
        \includegraphics[width=\textwidth]{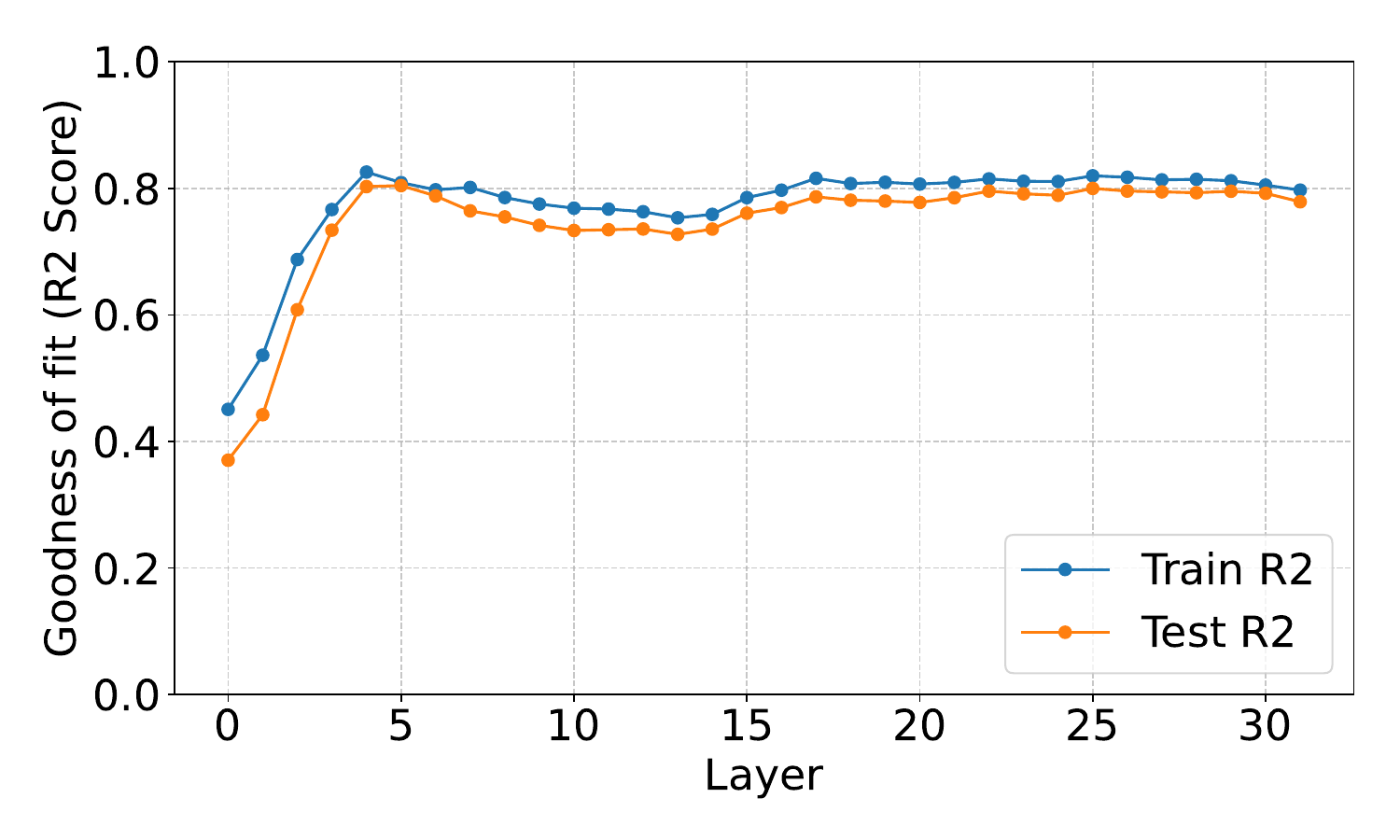}
        \caption{Latitude Attribute}
    \end{subfigure}
\caption{The $R^2$ score of predicting entity's numerical attributes, using a 5-Component PLS model.}
\label{fig:internal_repr_experiments1}
\end{figure*}

\begin{figure*}[ht!]
    \centering
    \begin{subfigure}[b]{0.32\textwidth}
        \includegraphics[width=\textwidth]{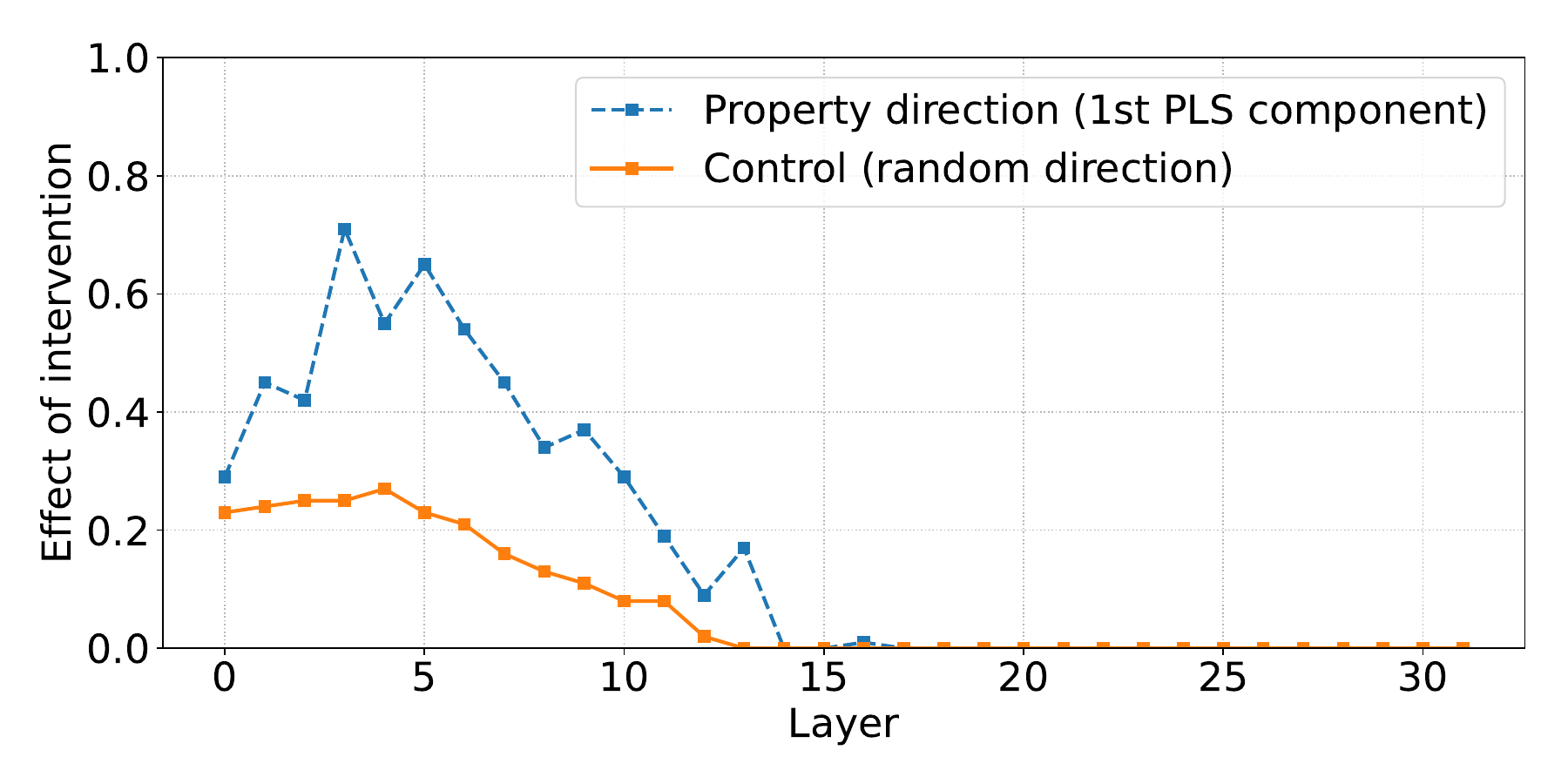}
        \caption{Birth Attribute}
    \end{subfigure}
    \hfill
    \begin{subfigure}[b]{0.32\textwidth}
        \includegraphics[width=\textwidth]{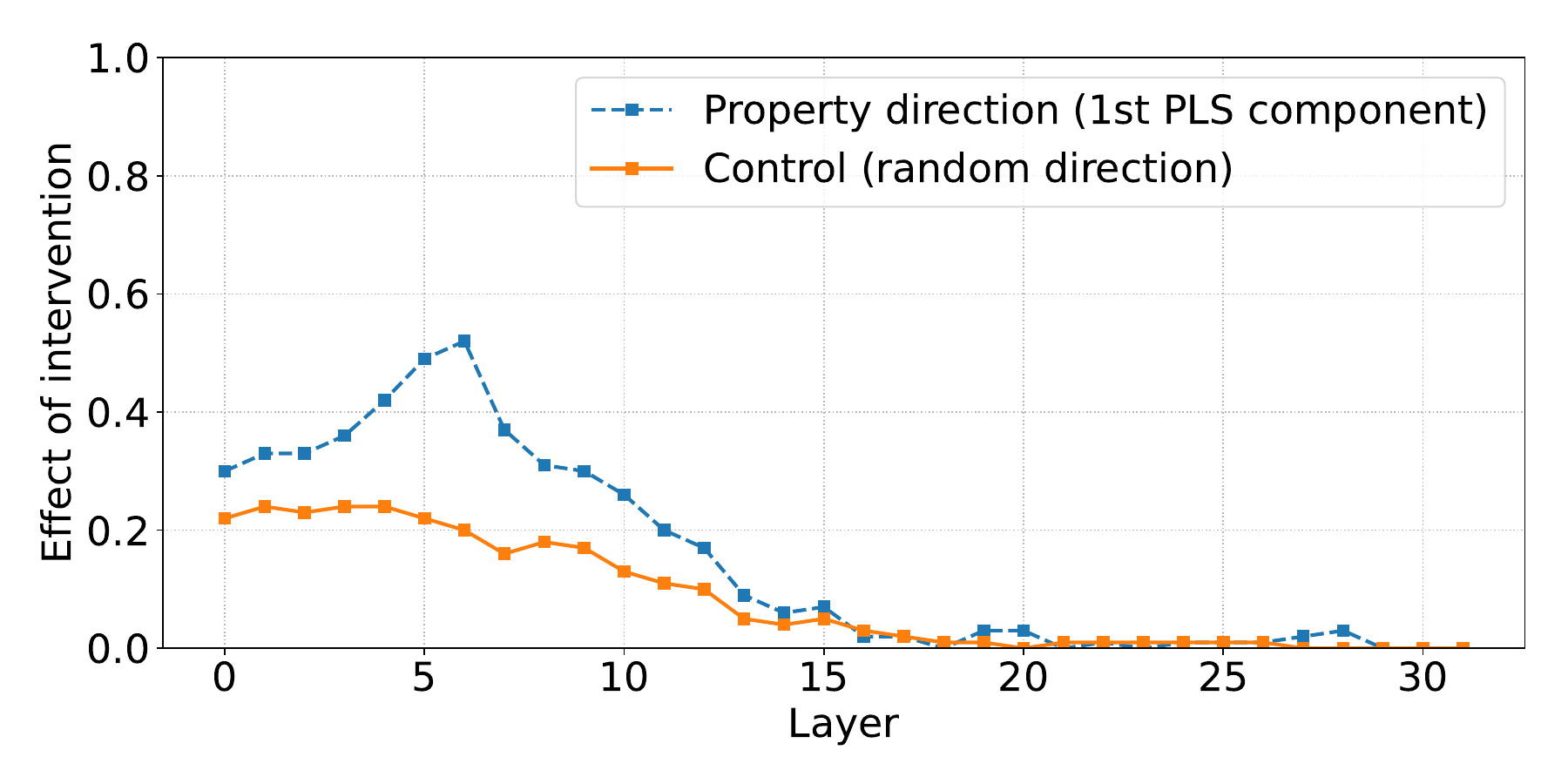}
        \caption{Death Attribute}
    \end{subfigure}
    \hfill
    \begin{subfigure}[b]{0.32 \textwidth}
        \includegraphics[width=\textwidth]{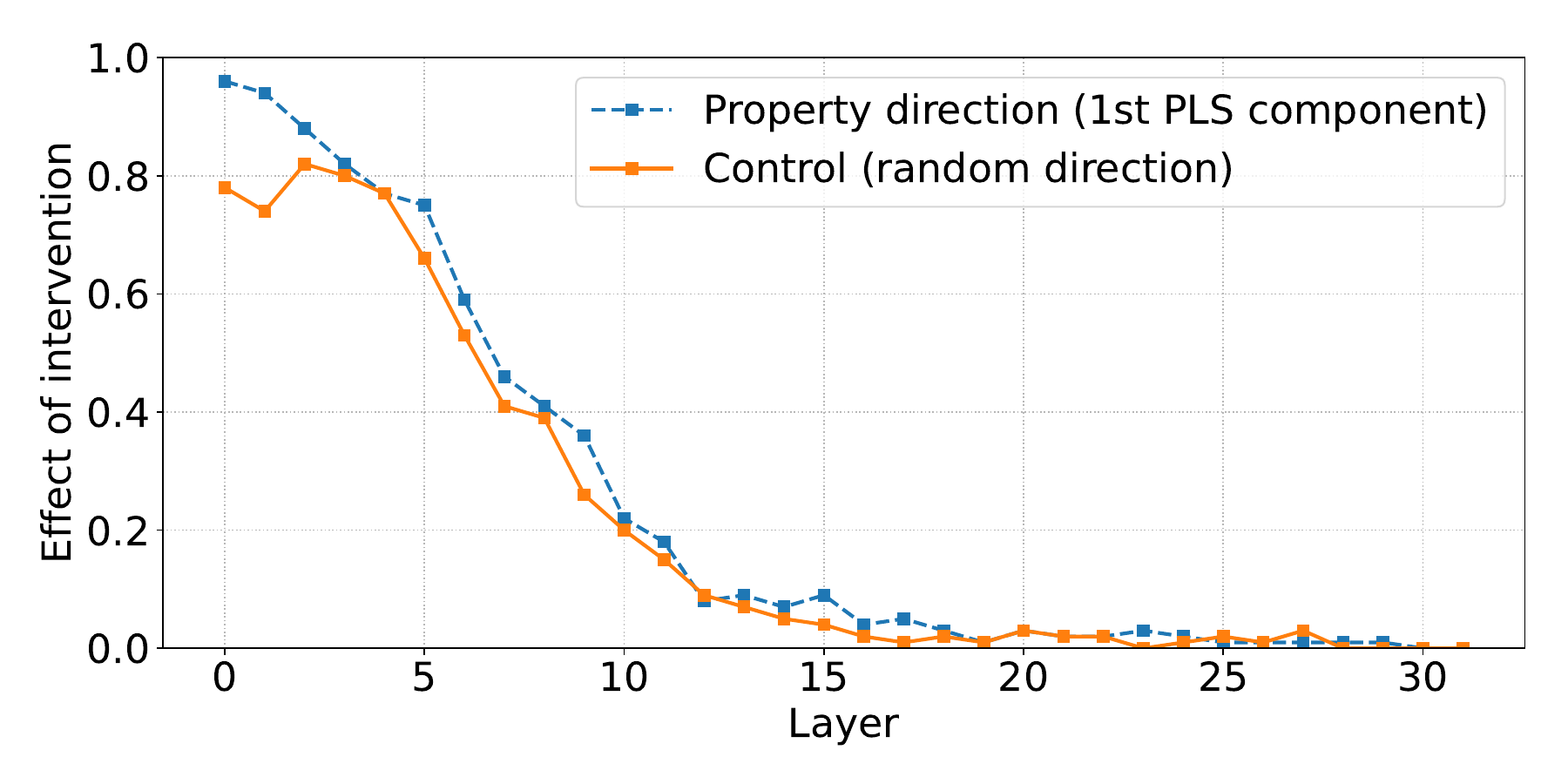}
        \caption{Latitude Attribute}
    \end{subfigure}

    \caption{The effect of the intervention—specifically, the ratio of flipped answers after performing intervention—was analyzed within the identified model subspace of each layer and compared to the effects observed in a randomly selected direction sampled from a normal distribution. }
    \label{fig:Intervention_results}
\end{figure*}

This experiment aims to train a PLS model to identify low-dimensional linear subspaces within the activation space, which could potentially be efficient in predicting numerical attributes for various entities. We then demonstrated the causal relationship within these subspaces by implementing targeted interventions which shows that indeed there is a causal effect between the identified linear subspaces and the logical comparison answers by the model. We validate our hypothesis by running three models on three numerical attributes. 

We also fitted another PLS model to evaluate Yes/No comparison reasoning related to these numerical attributes (see appendix ~\ref{internal_representation_exp2}).

\subsection{Prediction of numerical attributes with PLS}
\label{internal_representation_exp1}
 The training procedure consists of the following steps: (1) we first filter out the entities that the model predicted their comparison incorrectly (Section~\ref{behavioral_experiments_2}). (2) We feed a context vector that contains the comparison prompt (e.g., \textit{Was Cristiano born prior to Messi?)} (3) We extract the hidden states of the last token of each entity from the LLM's hidden states at a particular layer. (4) These hidden states are then used to train a PLS model with a $5$ component to predict the corresponding numerical attribute of each entity based on their corresponding model representation (activations). 
 Figure~\ref{fig:internal_repr_experiments1} depicts the results achieved by $five$ components PLS model, measured by the coefficient of determination $R^{2}$. The goodness of fit exceeds $0.8$ for all measured properties, indicating that the information encoded in these attributes can be extracted with low-dimensional (linear) subspaces. 

\begin{table}[t!]
\centering

\resizebox{\columnwidth}{!}{
\begin{tabular}{@{}l*{10}{r}@{}}
\toprule
& \multicolumn{10}{c}{Prompts} \\
\cmidrule(l{1pt}r{1pt}){2-11}
Task & 1 & 2 & 3 & 4 & 5 & 6 & 7 & 8 & 9 & 10 \\
\midrule
BP & $66.0$ & $70.0$ & $67.4$ & $66.2$ & $72.3$ & $67.6$ & $66.9$ & $66.6$ & $68.2$ & $71.3$ \\
DP & $63.4$ & $65.5$ & $61.5$ & $61.5$ & $67.0$ & $65.0$ & $63.3$ & $60.1$ & $61.7$ & $66.1$ \\
LP & $47.6$ & $72.0$ & $69.0$ & $70.0$ & $69.0$ & $68.5$ & $61.5$ & $69.0$ & $69.0$ & $66.6$ \\
\midrule
BC & $57.0$ & $56.6$ & $75.6$ & $67.0$ & $62.5$ & $50.0$ & $74.5$ & $57.0$ & $71.7$ & $62.1$ \\
DC & $53.5$ & $50.3$ & $74.8$ & $58.7$ & $50.5$ & $50.2$ & $50.3$ & $61.8$ & $50.1$ & $56.6$ \\
LC & $53.0$ & $56.0$ & $50.0$ & $37.8$ & $55.0$ & $51.2$ & $55.0$ & $50.0$ & $50.0$ & $50.2$ \\
\bottomrule
\end{tabular}

}
\caption{Experiments 1 and 2's Results for three tasks, and 10 different prompts for each. The accuracy of exact matching is reported, except for the Latitude task, where we relaxed the predicted and ground truth to be rounded to the integer part. \textbf{BP}: Birth Prediction, \textbf{DP}: Death Prediction, \textbf{LP}: Latitude Prediction, \textbf{BC}: Birth Comparison, \textbf{DC}: Death Comparison, \textbf{LC}: Latitude Comparison}
\label{table:behavioral experiments results}

\end{table}

\subsection{Intervention using PLS Components Vector}
\label{internal_analysis_exp3}

While the previous experiments with the PLS model establish correlation, they do not demonstrate causality. For this purpose, we perform interventions at a particular token within a designated model layer, chosen based on the correlation strength identified in predicting numerical attributes from each task (Section~\ref{internal_representation_exp1}).  We fix the first entity and intervene at the last token of the second entity. This token's hidden state is then updated by a scaled version of the first component direction from the PLS model to the original hidden state $h^{(l)}_i$ as illustrated in equation \eqref{equation1}.

In Figure~\ref{fig:Intervention_results} we compare the effect of our intervention per layer against a random vector from the normal distribution. It is measured by the Effect of Intervention metric (EI) (equation~\ref{equation: IE}), $f$ and $f'$ are the clean and patched models.   

\begin{equation}
   \text{EI} = \frac{1}{N}\sum_{i=1}^{N} \mathbb{I}\,[f(x_i) \neq f'(x_i)]
   \label{equation: IE}
\end{equation}
The results clearly demonstrate the superiority of our intervention method, particularly evident in Subfigures $a$ and $b$. In subfigure $c$, related to the Latitude numeric attribute, the gap between our method and the baseline narrows, suggesting that the direction may not be significant for this attribute. This could reflect the mode's nearly random response in the behavior experiment (Section~\ref{behavioral_experiments_2}). Additionally, the intervention's effect is notable only in the first $\approx50\%$ of the model layers, after which it diminishes to zero, aligned with prior research on inference time theory. 
We also tested the generalization of our approach on unseen samples, as shown in appendix, Figure~\ref{fig:Intervention_ood_results} and additional models (see \S~\ref{subsection: Experiments on Additional Models}). 

\begin{figure}
    \centering
    \includegraphics[width=0.5\textwidth]{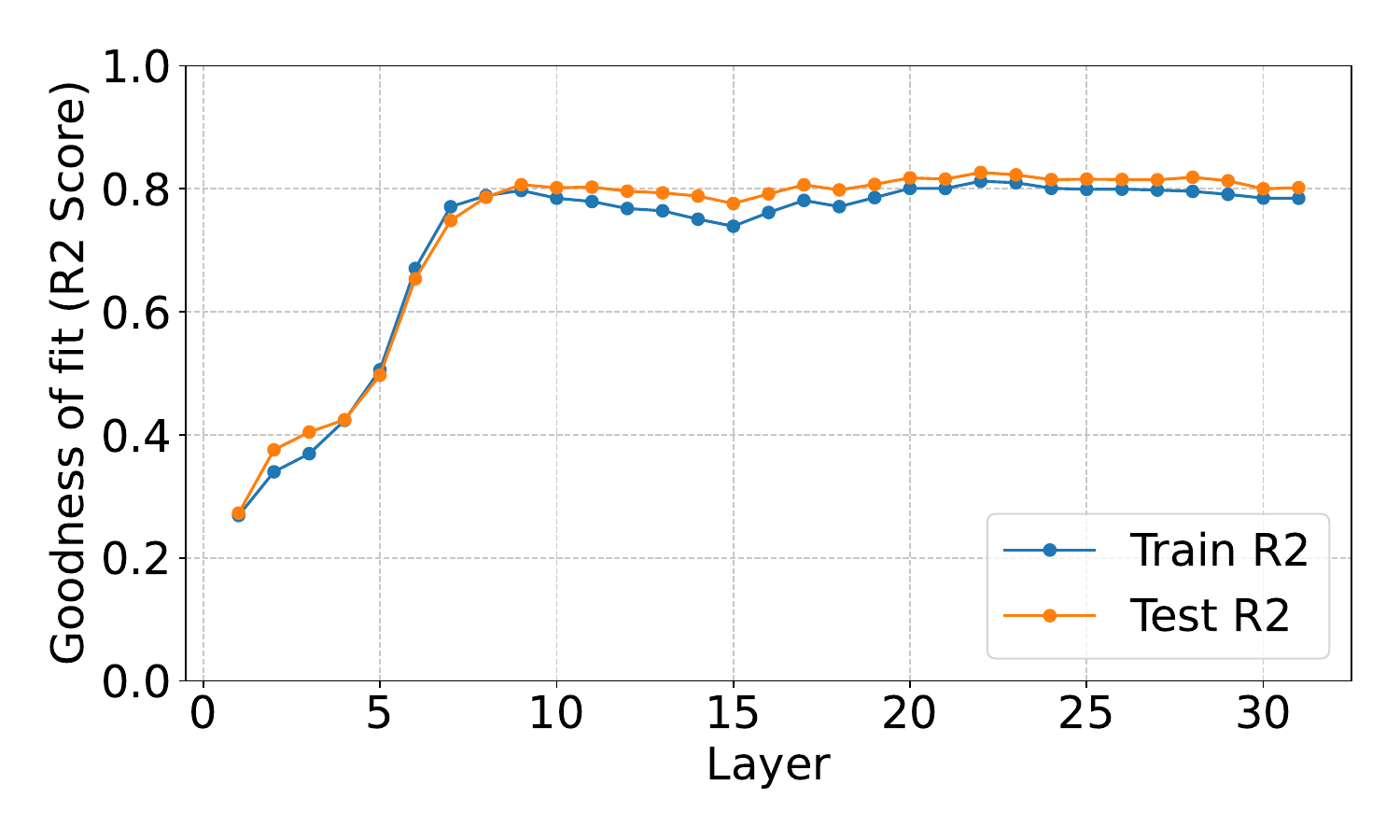}
    \caption{$R^2$ score of predicting entity's birth years attributes, using a 5-Component PLS model trained on Mistral 7B Instruct activations.}
    \label{fig:Mistral 7B birth r2 score}
\end{figure}

\begin{figure}
    \centering
    \includegraphics[width=0.5\textwidth]{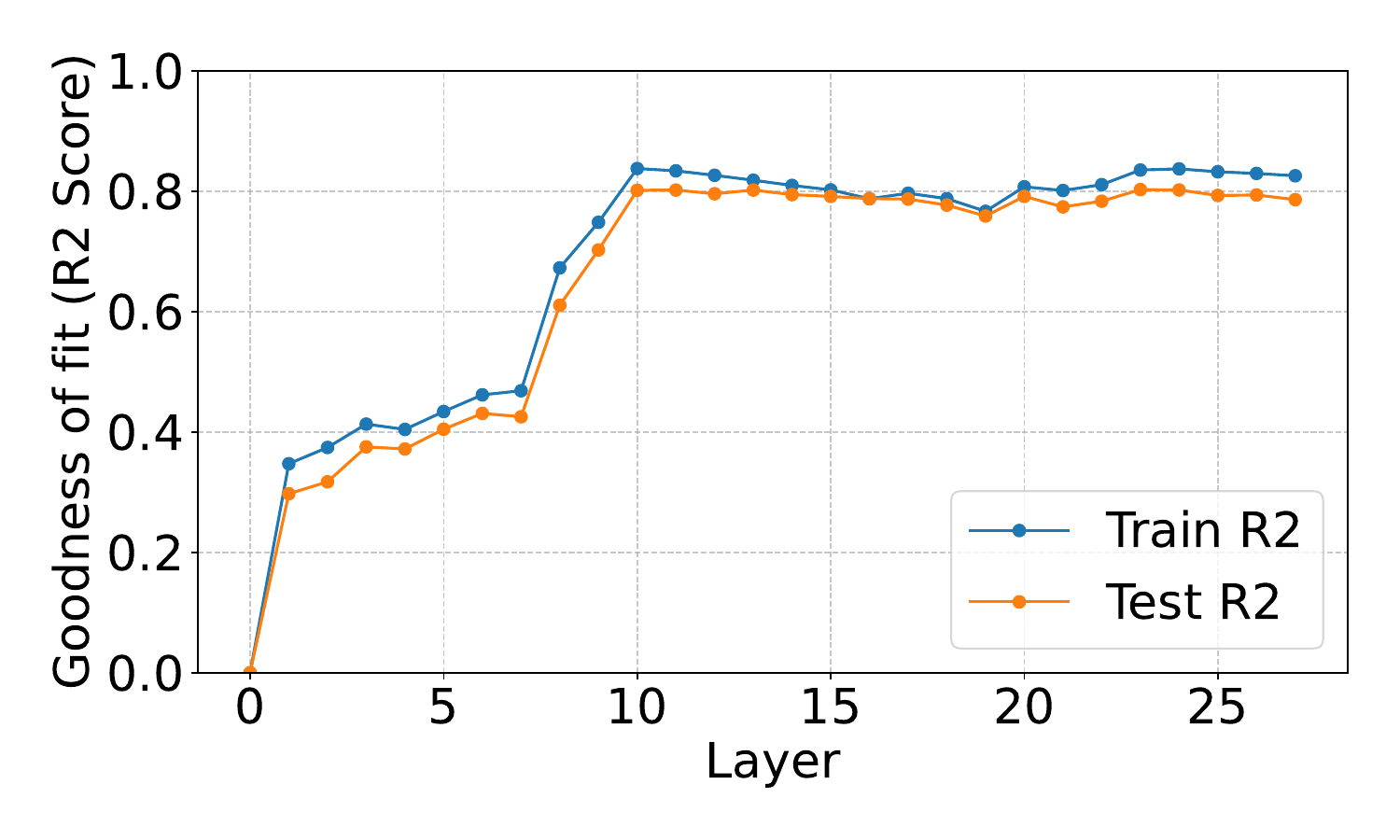}
    \caption{$R^2$ score of predicting entity's birth years attributes, using a 5-Component PLS model trained on Qwen2.5 7B Instruct activations.}
    \label{fig:Qwen2.5 7B Instruct birth r2 score}
\end{figure}

\begin{figure}
    \centering
    \includegraphics[width=0.5\textwidth]{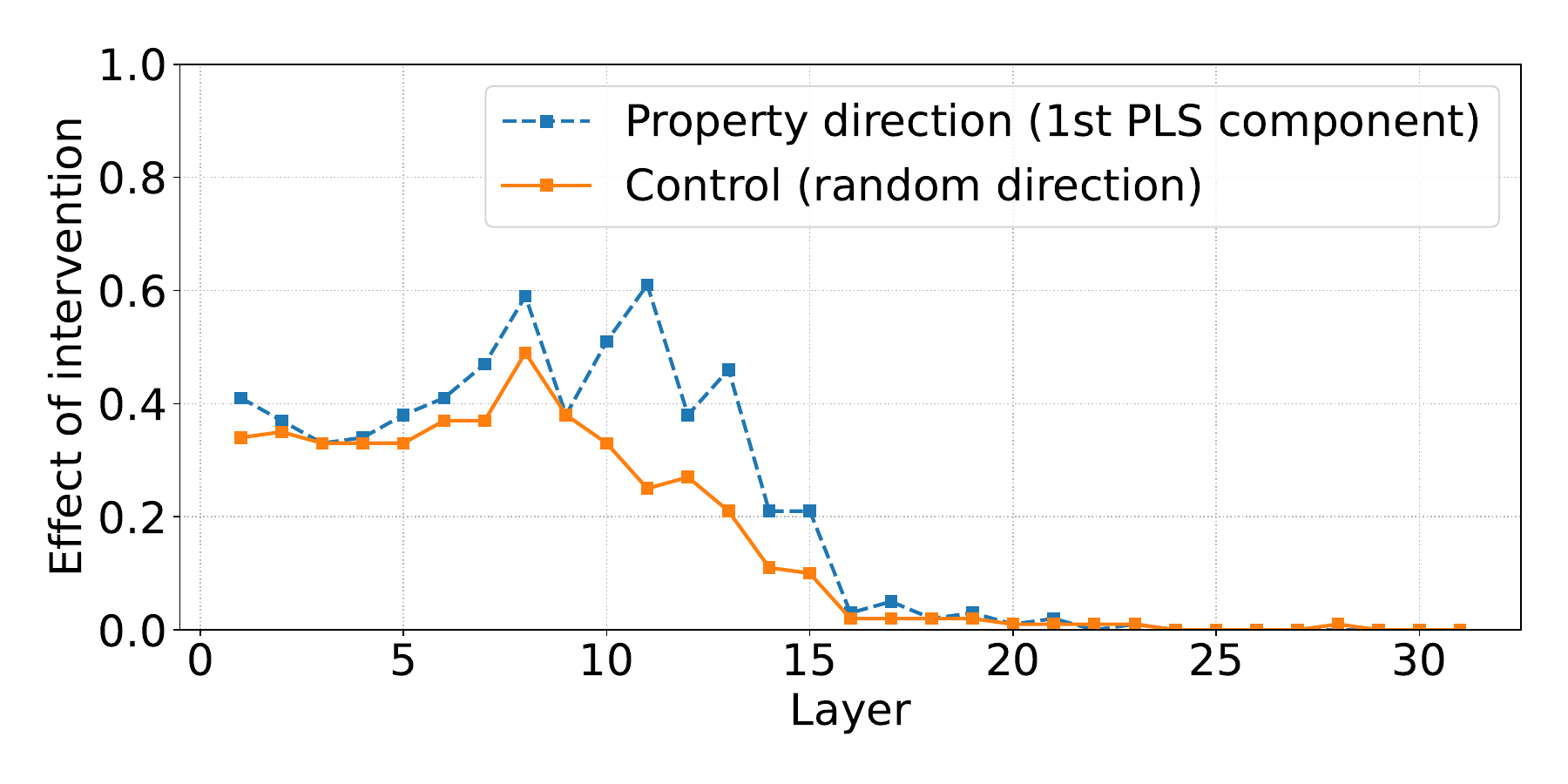}
    \caption{The effect of the intervention(i.e. the ratio of the flipped answers) in the identified subspace in each layer of the Mistral 7B Instruct model, compared to a random direction from a
normal distribution.}
    \label{fig:Mistral 7B birth intervention result}
\end{figure}

\begin{figure}
    \centering
    \includegraphics[width=0.5\textwidth]{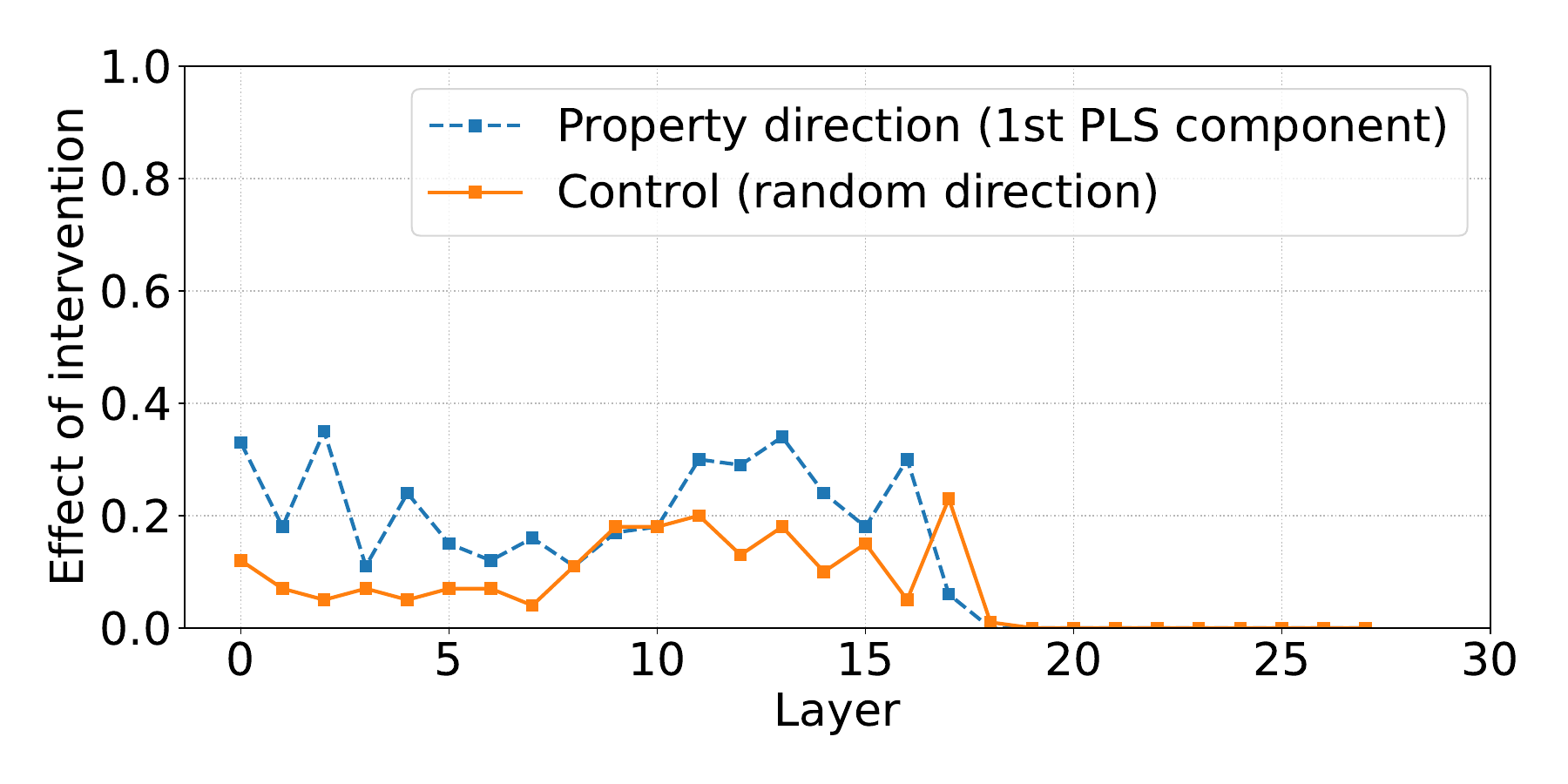}
    \caption{The effect of the intervention(i.e., the ratio of the flipped answer) in the identified subspace in each layer of the Qwen2.5 7B instruct model, compared to a random direction from a normal distribution.}
    \label{fig:Qwen 2 7B birth intervention result}
\end{figure}

\begin{table}[ht]
\centering

\label{table:combined_behavioral_experiments_results}
\resizebox{\columnwidth}{!}{
\begin{tabular}{@{}ll*{10}{r}@{}}
\toprule
Task & Model & \multicolumn{10}{c}{Prompts} \\
\cmidrule(lr){3-12}
 & & 1 & 2 & 3 & 4 & 5 & 6 & 7 & 8 & 9 & 10 \\
\midrule
\multirow{2}{*}{BP} 
 & Mistral 7B & 72.65 & 72.63 & 74.68 & 75.36 & 73.64 & 75.44 & 73.86 & 74.81 & 72.90 & 73.56 \\
 & Qwen2.5 7B & 40.82 & 34.68 & 33.95 & 34.59 & 33.95 & 36.72 & 36.96 & 32.61 & 39.07 & 34.32 \\
\cmidrule(lr){1-12}
\multirow{2}{*}{BC} 
 & Mistral 7B & 53.60 & 64.84 & 64.02 & 53.10 & 61.88 & 57.66 & 53.00 & 67.06 & 64.68 & 50.00 \\
 & Qwen2.5 7B & 29.20 & 58.10 & 38.88 & 26.22 & 49.76 & 40.54 & 6.20 & 3.84 & 9.16 & 6.98 \\
\bottomrule
\end{tabular}
}
\caption{Exact Matching Accuracy of Mistral 7B and Qwen2.5 7B Models on Birth Date Numerical property extraction and Comparison Tasks Across Prompt Variations. All models are instruction-based models. \textbf{BP:} Birth Prediction and \textbf{BC:} Birth Comparison tasks are evaluated.}
\end{table}

\subsection{Experiments on Additional Models}
\label{subsection: Experiments on Additional Models}
To further validate our hypothesis generalization, we run the same experiments on two additional language models for the \textit{birth} property. Those additional models are Mistral-7B-intruct~\cite{jiang2023mistral7b} and Qwen2.5-7B-Instruct~\cite{qwen2.5}. 

PLS models trained on models' activation have crossed an $R^2$ score of $0.8$ suggesting that the information encoded in those models' activations can be extracted using low-dimensional (linear) subspaces (see Figure~\ref{fig:Mistral 7B birth r2 score} and Figure~\ref{fig:Qwen2.5 7B Instruct birth r2 score}).   

The Effect of Intervention (EI) results shown in Figures~\ref{fig:Mistral 7B birth intervention result} and ~\ref{fig:Qwen 2 7B birth intervention result} of the Mistral 7B Instruct and Qwen-2.5 7B Instruct models, respectively, demonstrate the same behavior seen in the previous experiments. For the EI of Mistral, we can see that the peak was around the $11$th layer and then continued to decrease until it finally disappeared around the $16$ layer (Figure~\ref{fig:Mistral 7B birth intervention result}). When compared to other models, Qwen2.5 has shown two clear differences. First, we can observe two peaks for the EI with almost the same value of the EI, early around the $third$ layer and later one around layer $12$, while other models have shown only one peak. Second, unlike other models, Qwen2.5 7B kept bouncing around almost the same EI values and suddenly become None at around layer $16$ (Figure~\ref{fig:Qwen 2 7B birth intervention result}). One reason that might explain the difference between Qwen2.5 7B and other models, is that Qwen2.5 7B uses only $28$ layers, while other models in the experiments are formed of $32$ layers.





\section{Conclusion}
In this research, we empirically demonstrate that the model answers numerical reasoning questions, such as "Was Cristiano born before Messi?" using a two-step process. First, it extracts numerical attributes for each entity from a linear subspace. The second step involves utilizing these linear directions to answer the logical question. Specifically, subspaces are identified through PLS regression, where directions in low-dimensional subspaces of the activation space encode numerical property information. We illustrate this approach using three numerical attributes: Birth, Death, and Latitude across three LLMs. The reasoning step is validated using causal interventions along the direction of the first component of the PLS model, where these interventions successfully alter the model's answers.

\section{Ethical Statement}
Our work adheres to the ACL Code of Ethics and maintains a high standard of ethical research practice. We ensure that our methodology, data usage, and model development follow responsible AI principles, and that there are no ethical violations in our study. Our research does not involve the use of sensitive or private data, nor does it contribute to any potential harm or bias propagation. We remain committed to transparency, fairness, and the responsible application of large language models in line with ACL’s ethical guidelines.

\section{Limitations}
\label{section:limitations}
This work has several limitations we plan to address in future work:
\begin{itemize}
    \item Error Analysis: While the experimental results demonstrate the model's ability to map numerical properties to low-dimensional subspaces and use them for reasoning tasks, we have not conducted a thorough error analysis to understand the model's types of mistakes. Identifying patterns in erroneous outputs could guide improvements in both model design and training.


\item Limited Scope of Numerical Attributes: Our experiments are restricted to three types of numerical attributes: birth year, death year, and geographic latitude. It remains unclear whether our findings extend to a broader range of numerical properties, such as financial data, time intervals, or other continuous variables. We plan to investigate this in future work.

\item Intervention Hyperparameter Sensitivity: The success of the intervention experiments relies heavily on the choice of the scaling factor $\alpha$ applied during the intervention. We have not explored the full sensitivity of the model’s performance to this hyperparameter, which could introduce biases or instability in real-world applications.

\end{itemize}

\bibliography{custom}

\appendix

\section{Background}
\label{section: Background}
\paragraph{Generative-Transformer Language Models.} 
Transformer models, particularly in generative contexts, have revolutionized natural language processing tasks due to their self-attention mechanisms. These models map an input sequence $x_1, x_2, \dots, x_n$ to a corresponding sequence $y_1, y_2, \dots, y_m$ using multi-layer perceptron, and multi-head self-attention layers, which compute attention scores based on the query-key-value system. Mathematically, for a given layer $l$, the attention output $A_l$ is computed as:

\begin{equation}
    A_l = \text{softmax} \left( \frac{Q K^T}{\sqrt{d_k}} \right) V
\end{equation}

where $Q$, $K$, and $V$ are the query, key, and value matrices, and $d_k$ is the dimension of the keys. By stacking multiple layers of these attention mechanisms and multi layer percptron, transformers efficiently capture long-range dependencies in text. The autoregressive nature of generative transformers allows them to generate coherent text sequences by predicting the next token based on previous tokens. 

\paragraph{Representation Analysis of Transformer Language Models.} 
Representation analysis of transformers has revealed important insights into how these models store and manipulate information across layers. Research has shown that transformer language models develop complex, hierarchical representations that can be understood by analyzing the attention patterns and hidden states at different layers \citep{niu2024does}. For example, studies have found that early layers capture syntactic structures, while deeper layers capture more semantic information \citep{hernandez2023linearity}. Recent work also uses probing techniques to analyze how specific linguistic features are represented, contributing to a growing understanding of model interpretability \cite{vulic2020probing}.

\paragraph{Intervention and Activation Patching.} One technique that has gained attention in the analysis of neural models, including transformers, is \textbf{activation patching}. This involves replacing activations in a specific layer with those from another input in order to study the effect of those activations on the final output. By intervening at different points within the model, researchers can better understand how information is processed and transformed throughout the network. This method has been useful in dissecting how specific neurons or attention heads contribute to a model's behavior, allowing for targeted interventions that shed light on model interpretability.

\paragraph{Linear Hypothesis in Representation.}The \textbf{linear hypothesis} posits that the representations formed by transformer models are linearly separable. This means that complex patterns, such as syntactic and semantic categories, can be distinguished by applying a linear transformation to the learned embeddings \citep{park2023linear}. The key idea here is that the hidden representations of different tasks or features align in such a way that linear classifiers can achieve good performance with minimal processing, a phenomenon observed across a range of neural architectures. Connecting this with the previous analysis, it appears that transformers structure their internal space in a way that is amenable to linear separation of features, thus facilitating tasks such as classification and regression.

\paragraph{Partial Least Squares (PLS).}
Partial Least Squares (PLS) offers an alternative to Principal Component Analysis (PCA) for dimensionality reduction, especially when predicting one set of variables from another. PLS seeks to maximize the covariance between the input matrix $\mathbf{X}$ and the response matrix $\mathbf{Y} $ by projecting both onto a latent space. The key idea is to find latent variables $\mathbf{T} = \mathbf{XW}$ and $\mathbf{U} = \mathbf{YC}$ that best capture this covariance. 

The predictive relationship between $\mathbf{X}$ and $\mathbf{Y}$ is then modeled as:
\begin{equation}
\label{eq1}
    \hat{\mathbf{Y}} = \mathbf{X} \mathbf{W} \mathbf{P}^T,
\end{equation}
where $\hat{\mathbf{Y}}$ is the predicted output matrix, $\mathbf{P}$ are the loadings, and the quality of this prediction can be assessed using the coefficient of determination \(R^2\). The \(R^2\) value measures how well the model explains the variance in $\mathbf{Y}$, where higher values indicate a better fit between predicted and actual outputs.

PLS is preferred over regression when predictors (or columns of  $\mathbf{X}$) are not independent or when the number of predictors exceeds the number of observations, making it suitable for high-dimensional data. For transformers, applying PLS helps uncover how input embeddings influence predictions by focusing on the shared variance between input features and outputs \citep{heinzerling2024monotonicrepresentationnumericproperties}.






\section{Related Work}
\label{section: Related word}
After the appearing of pre-trained language models such as ELMo~\citep{peters2018deep}, BERT~\citep{devlin2019bert}, and GPT~\cite{Radford2018ImprovingLU}, researchers have had interests in the numerical capability of language models.
\cite{spithourakis2018numeracy} evaluates the pre-trained language models from viewpoints of the output capability of numerical tokens, the behavioural side of the numeracy.
\cite{wallace2019nlp} focused on the numerical knowledge stored in the embeddings, which is the internal side of the numeracy.
\newcite{zhang2024interpreting} investigated the internal working of the recent large language models when processing arithmetic calculation.

Knowledge of entities such as named entity has also been payed attention to by many researchers.
Considering the pre-trained language models as a knowledge base~\cite{petroni2019language,jiang2020how}, behavioral~\cite{shin2020autoprompt} and internal~\cite{meng2022locating,dai2022knowledge} analysis have been studied.

With much larger scale of language models such as GPT3~\cite{brown2020language} and LLaMA~\cite{touvron2023llama2} and the technique of in-context learning, the capability of reasoning acquired by the language models has started to be discussed.
\cite{merullo2024a} examined the internal working of language models when solving the reasoning task of the entity-entity relation such as \textit{(Paris, capital-of, France)}.
\newcite{heinzerling2024monotonicrepresentationnumericproperties} provides a deeper observation of the reasoning of the entity-numeric relation such as \textit{(Dijkstra, born-in, 1930)}.
They reveal that the entity-numeric relations are stored in the language models' representation as keeping their monotonic structure.
Following this work, we further dive into the numerical reasoning that requires the extraction of the entity-numeric knowledge and the comparison of the two numerical information such as \textit{(Bellman, born-before, Dijkstra)}.

\subsection{Logical Comparison with PLS}
\label{internal_representation_exp2}

\begin{figure*}[ht!]
    \centering

    \begin{subfigure}[b]{0.32\textwidth}
        \includegraphics[width=\textwidth]{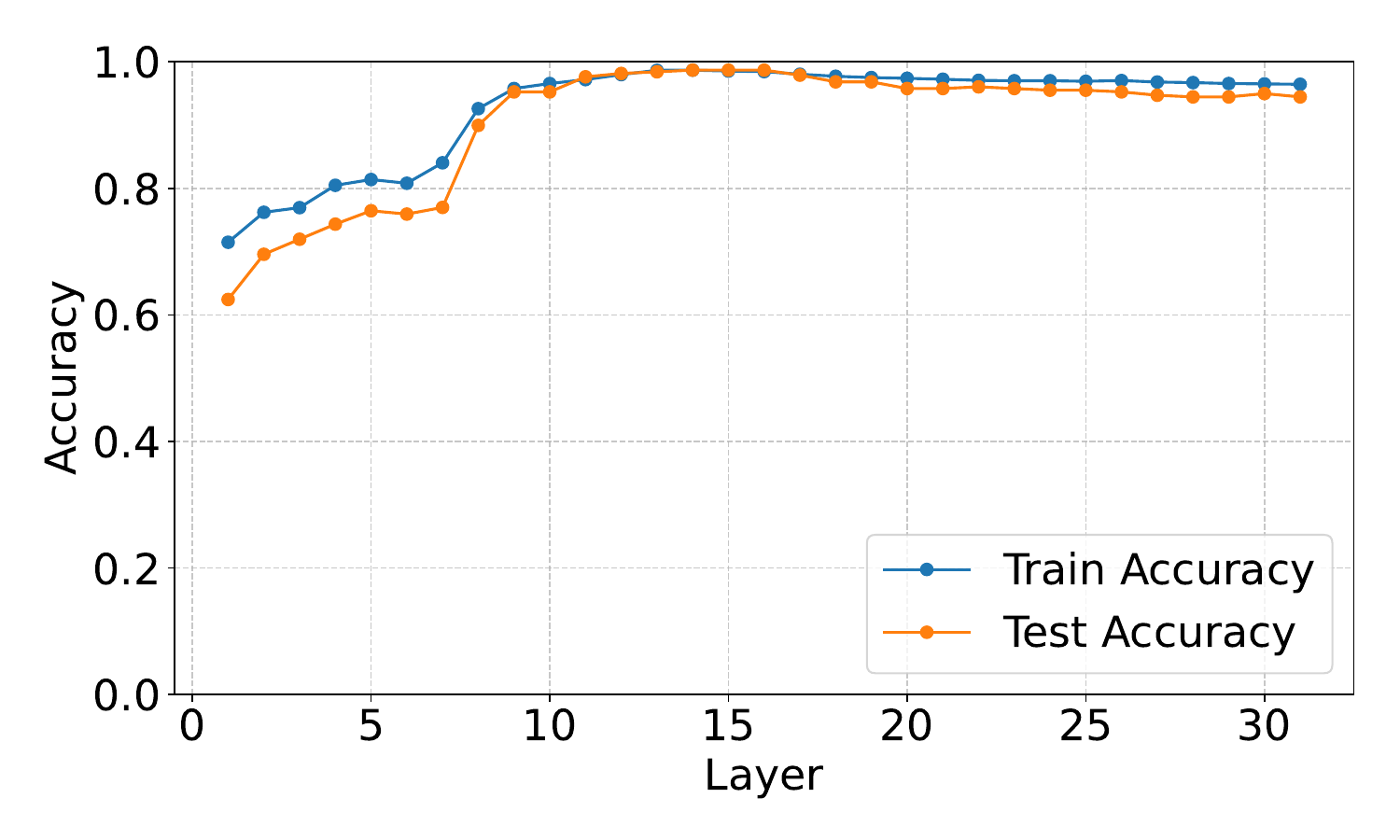}
        \caption{Birth Attribute}
    \end{subfigure}
    \hfill
    \begin{subfigure}[b]{0.32\textwidth}
        \includegraphics[width=\textwidth]{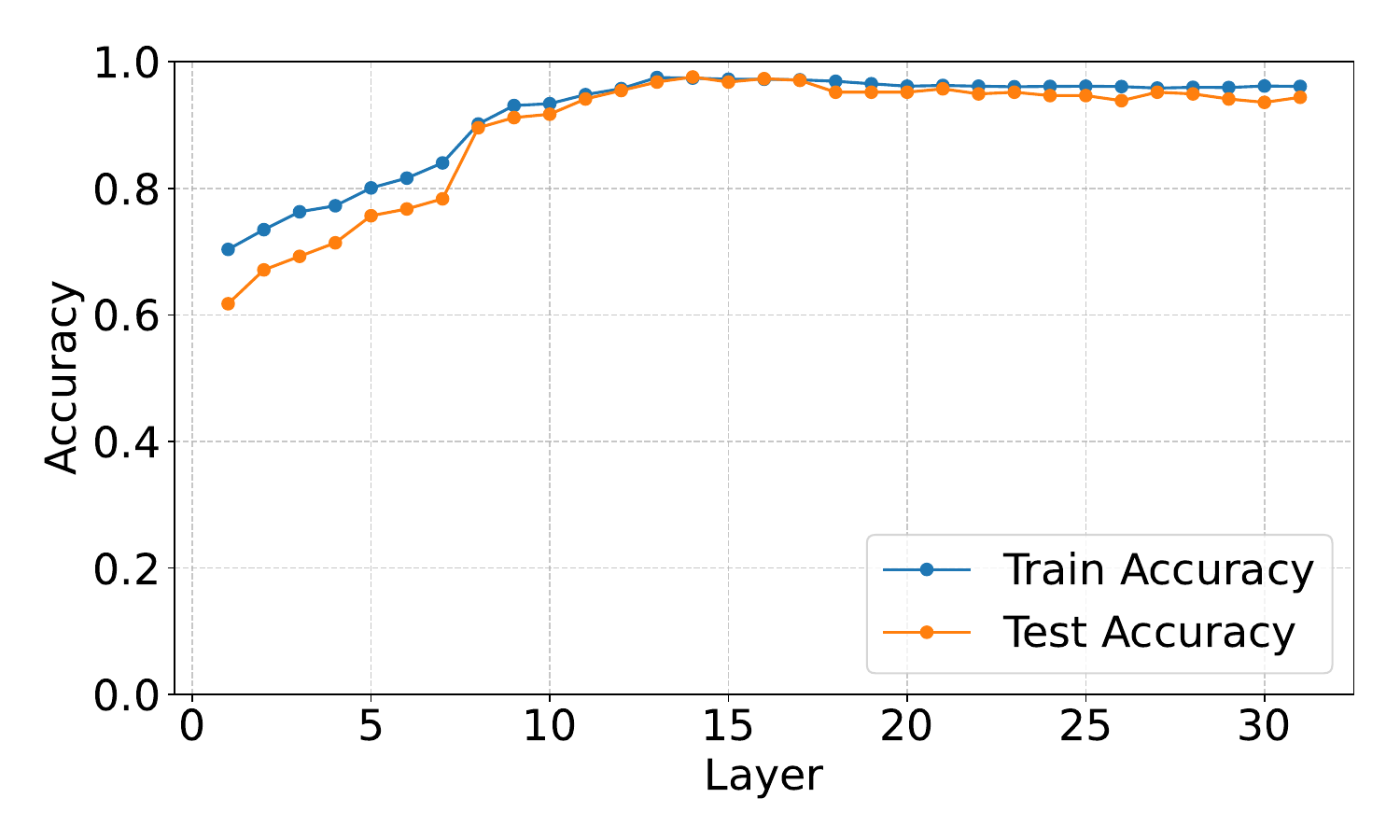}
        \caption{Death Attribute}
    \end{subfigure}
    \hfill
    \begin{subfigure}[b]{0.32\textwidth}
        \includegraphics[width=\textwidth]{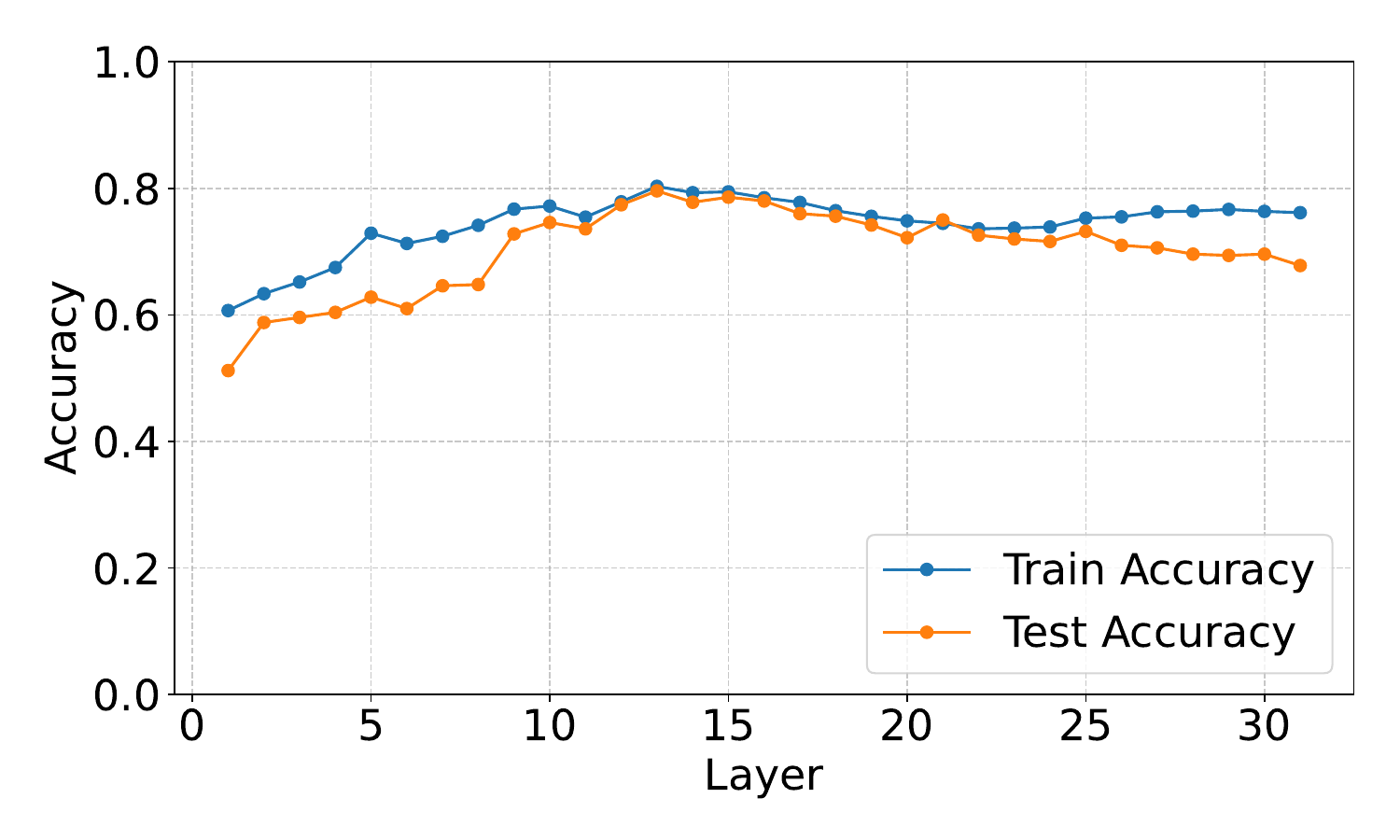}
        \caption{Latitude Attribute}
    \end{subfigure}

    \caption{The accuracy of predicting Yes/No in a comparison task of numerical attributes, using a 5-Component PLS model.}
\label{fig:internal_repr_experiments2}
\end{figure*}

In this experiment, we feed the entire context vector containing a comparison into the model and extract the last hidden state of the last token for each comparison sample. We train a PLS model on these activations to predict the comparison results (i.e. Yes or No). We aim to make sure that the Yes/No task is predictable from model activations using a low-dimensional (linear) subspace. Figure~\ref{fig:internal_repr_experiments2} illustrates the accuracy of the $5$-components PLS model in predicting the comparison results giving the model activations. The model shows near-perfect performance of the Birth and Death tasks, while less robust on the Latitude task. This outcome is consistent with findings from the Behavioral experiments in Section~\ref{behavioral_experiments_2}.

\begin{figure*}[ht!]
    \centering
    \begin{subfigure}[b]{0.32\textwidth}
        \includegraphics[width=\textwidth]{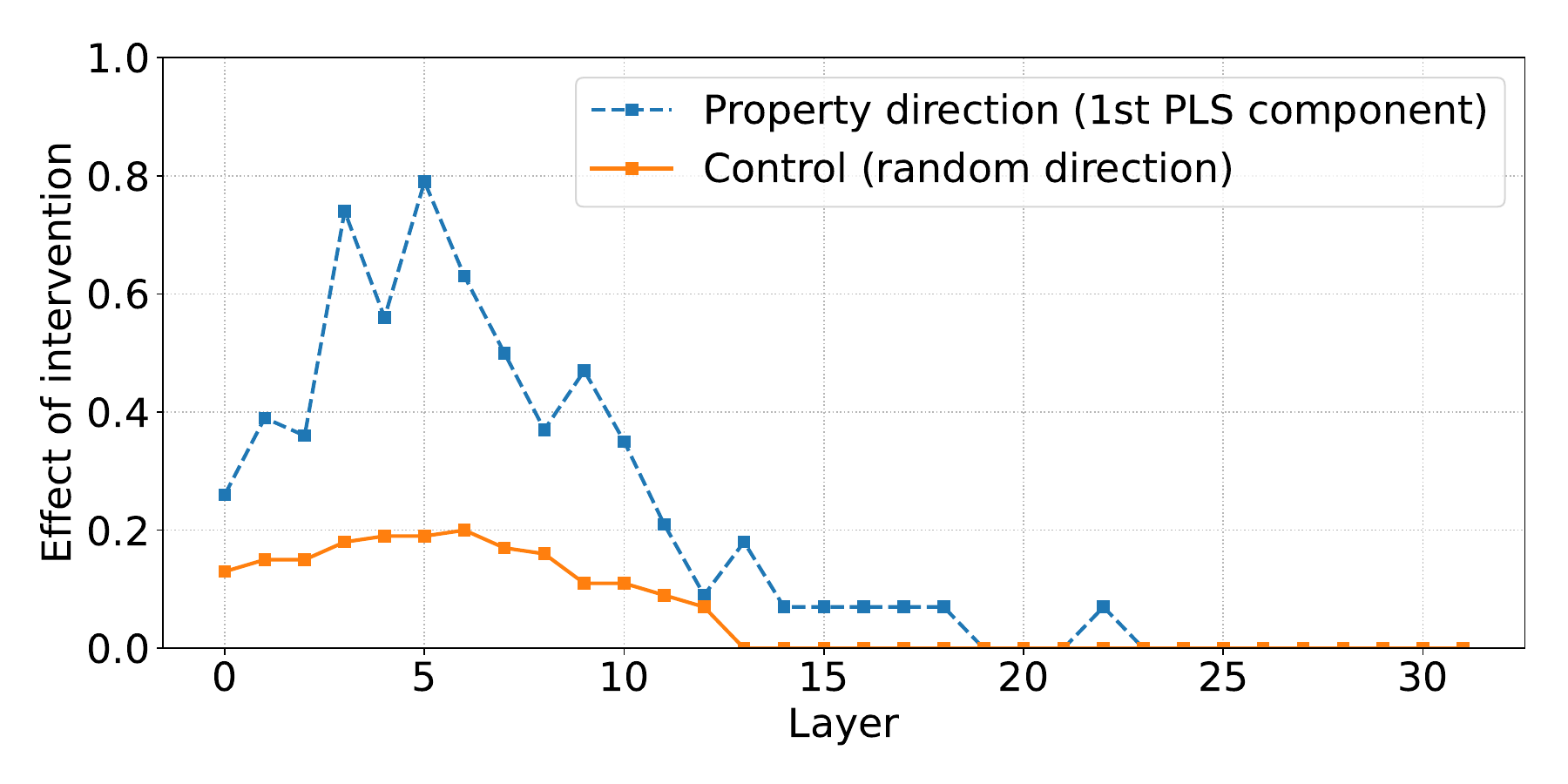}
        \caption{Birth intervention}
    \end{subfigure}
    \hfill
    \begin{subfigure}[b]{0.32\textwidth}
        \includegraphics[width=\textwidth]{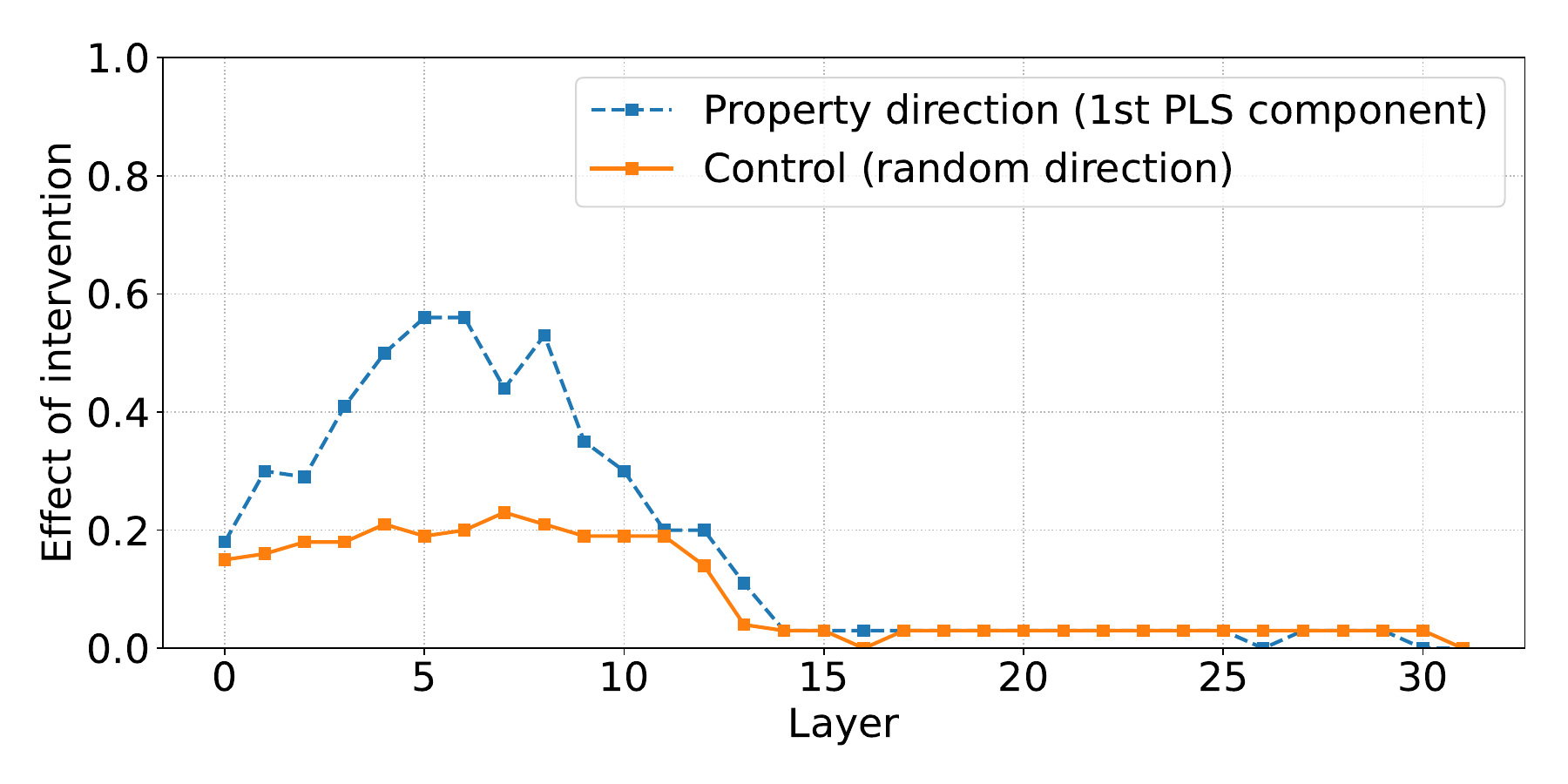}
        \caption{Death intervention}
    \end{subfigure}
    \hfill
    \begin{subfigure}[b]{0.32\textwidth}
        \includegraphics[width=\textwidth]{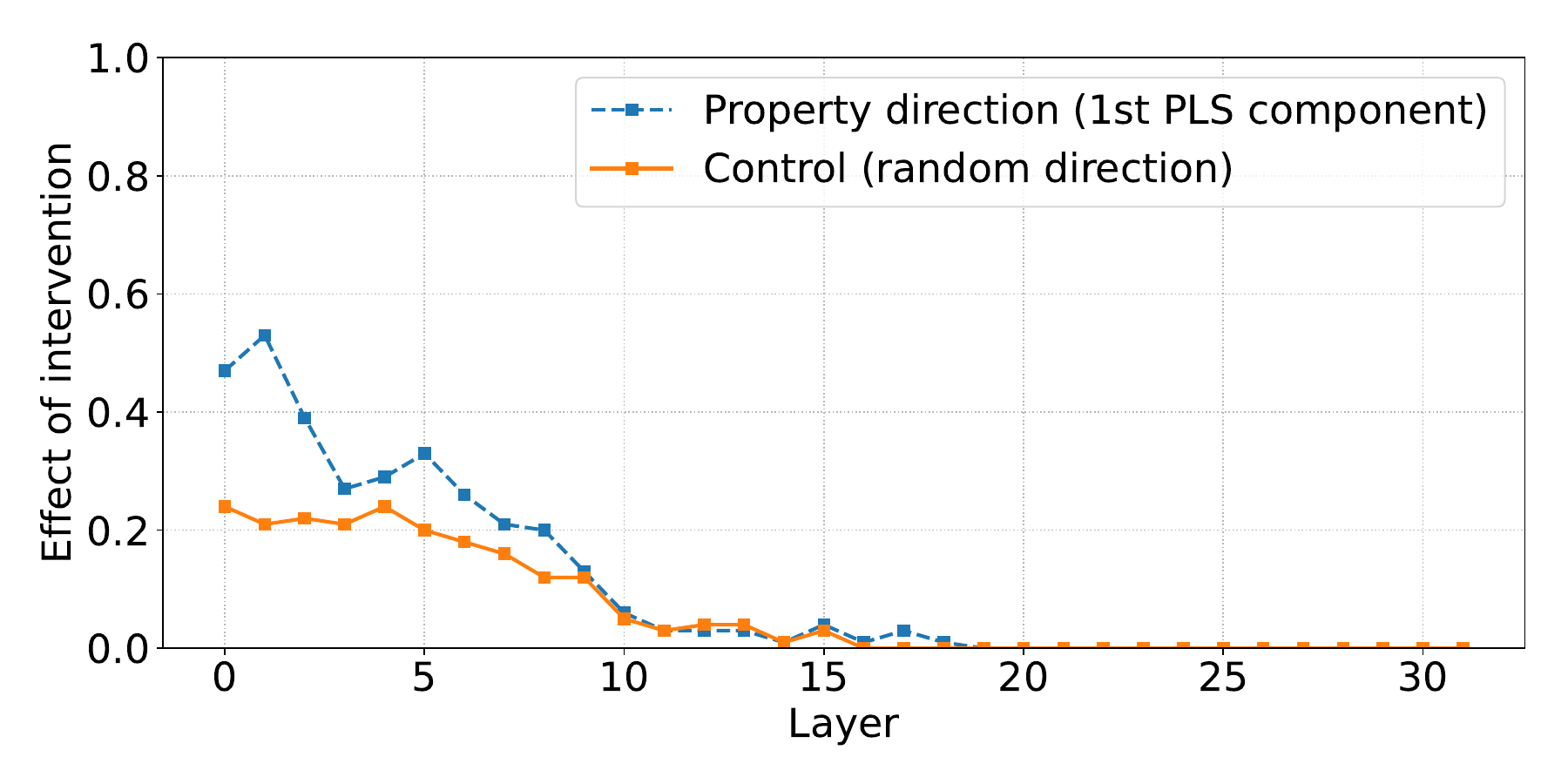}
        \caption{Latitude}
    \end{subfigure}

    \caption{Intervention graphs for out-of-distribution data samples on birth, death, and latitude tasks.}
    \label{fig:Intervention_ood_results}
\end{figure*}

\begin{table*}[ht]
\centering
\begin{tabular}{|p{0.3\textwidth}|p{0.3\textwidth}|p{0.3\textwidth}|}
\hline
\textbf{Birth} & \textbf{Death} & \textbf{Latitude} \\
\hline
Did \{entity\_x\} come into the world earlier than \{entity\_y\}? Answer with Yes or No. & Did \{entity\_x\} die before \{entity\_y\}? Answer with Yes or No. & Is \{entity\_x\} located at a higher latitude than \{entity\_y\}? Answer Yes or No. \\
\hline
Is \{entity\_x\}'s birthdate before \{entity\_y\}'s? Respond with Yes or No. & Did \{entity\_x\} pass away earlier than \{entity\_y\}? Respond with Yes or No. & Is \{entity\_x\} farther north than \{entity\_y\}? Answer Yes or No. \\
\hline
Was \{entity\_x\} born prior to \{entity\_y\}? Output only Yes or No. & Was \{entity\_x\}'s death prior to \{entity\_y\}? Provide only Yes or No. & Does \{entity\_x\} have a higher latitude value than \{entity\_y\}? Answer Yes or No. \\
\hline
Did \{entity\_x\} enter life before \{entity\_y\}? Answer with Yes or No. & Did \{entity\_x\} pass on before \{entity\_y\}? Answer Yes or No. & Comparing latitudes, is \{entity\_x\} north of \{entity\_y\}? Answer Yes or No. \\
\hline
Was \{entity\_x\}'s birth earlier than \{entity\_y\}'s? Output only Yes or No. & Did \{entity\_x\} die first compared to \{entity\_y\}? Respond only with Yes or No. & In terms of latitude, is \{entity\_x\} above \{entity\_y\}? Answer Yes or No. \\
\hline
Was \{entity\_x\} born first compared to \{entity\_y\}? Respond with Yes or No. & Was \{entity\_x\}'s death earlier than \{entity\_y\}'s? Answer with Yes or No. & Is the latitude of \{entity\_x\} greater than the latitude of \{entity\_y\}? Answer Yes or No. \\
\hline
Is \{entity\_x\} older than \{entity\_y\}? Reply only with True or False. & Did \{entity\_x\} precede \{entity\_y\} in death? Reply only with True or False. & Geographically, is \{entity\_x\} at a more northern latitude than \{entity\_y\}? Answer Yes or No. \\
\hline
Did \{entity\_x\} precede \{entity\_y\} in birth? Respond only with True or False. & Did \{entity\_x\} pass before \{entity\_y\}? Respond only with True or False. & Does \{entity\_x\} have a more northerly latitude compared to \{entity\_y\}? Answer Yes or No. \\
\hline
Did \{entity\_x\} arrive before \{entity\_y\}? Answer only with True or False. & Did \{entity\_x\} die earlier than \{entity\_y\}? Answer only with Yes or No. & Is \{entity\_x\} positioned at a latitude north of \{entity\_y\}? Answer Yes or No. \\
\hline
Is \{entity\_x\} senior to \{entity\_y\}? Reply only with Correct or Incorrect. & Did \{entity\_x\} pass away first compared to \{entity\_y\}? Reply with Correct or Incorrect. & Considering only latitude, is \{entity\_x\} more northward than \{entity\_y\}? Answer Yes or No. \\
\hline
\end{tabular}
\caption{Comprehensive list of prompts for our three tasks:  for Birth, Death, and Latitude}
\label{tab:comprehensive_list_of_prompts}
\end{table*}

\end{document}